\documentclass[letterpaper, 10 pt, conference]{ieeeconf}  

\pdfoutput=1
\usepackage{cite}
\usepackage{amsmath,amssymb,amsfonts}
\usepackage{algorithmic}
\usepackage{graphicx}
\usepackage{textcomp}
\usepackage{url}
\usepackage{xcolor}
\usepackage{subcaption} 
\usepackage{fancyhdr}
\usepackage[utf8]{inputenc}
\def\BibTeX{{\rm B\kern-.05em{\sc i\kern-.025em b}\kern-.08em
    T\kern-.1667em\lower.7ex\hbox{E}\kern-.125emX}}

\IEEEoverridecommandlockouts                              

\title{\LARGE \bf
Learning Humanoid Robot Running Skills through Proximal Policy Optimization
}

\author{Luckeciano C. Melo$^{1}$ and Marcos R. O. A. M\'{a}ximo$^{1}$
\thanks{$^{1}$Luckeciano Melo and Marcos M\'{a}ximo are with the Autonomous Computational Systems Lab (LAB-SCA), Computer Science Division, Aeronautics Institute of Technology, Praça Marechal Eduardo Gomes, 50, Vila das Acácias, 12228-900, São José dos Campos, SP, Brazil
        {\tt\small luckeciano@gmail.com, mmaximo@ita.br}}%
}

\begin{document}

\maketitle

\begin{abstract}
In the current level of evolution of Soccer 3D, motion control is a key factor in team’s performance. Recent  works  takes  advantages  of  model-free  approaches based  on  Machine  Learning to exploit  robot  dynamics in  order  to  obtain  faster locomotion skills, achieving running policies and, therefore, opening a new research direction in the Soccer 3D environment.

In this work, we present a methodology based on Deep Reinforcement Learning that learns running skills without any prior knowledge, using a neural network whose inputs are related to robot's dynamics. Our results outperformed the previous state-of-the-art sprint velocity reported in Soccer 3D literature by a significant margin. It also demonstrated improvement in sample efficiency, being able to learn how to run in just few hours.

We reported our results analyzing the training procedure and also evaluating the policies in terms of speed, reliability and human similarity. Finally, we presented key factors that lead us to improve previous results and shared some ideas for future work.

\end{abstract}

\section{Introduction}

RoboCup 3D Soccer Simulation League (Soccer 3D) is a particularly interesting challenge concerning humanoid robot soccer. It consists of a simulation environment of a soccer match with two teams, each one composed by up to 11 simulated NAO robots \cite{gouaillier2009}, the official robot used for RoboCup Standard Platform League since 2008. Soccer 3D is interesting for robotics research since it involves high level multi-agent cooperative decision making while providing a physically realistic environment which requires control and signal processing techniques for robust low level skills \cite{DBLP:journals/corr/abs-1901-00270}.

In the current level of evolution of Soccer 3D, motion control is a key factor in team's performance. Indeed, controlling a high degrees of freedom humanoid robot is acknowledged as one of the hardest problems in Robotics. Much effort has been devised to humanoid robot walking, where researchers have been very successful in designing control algorithms which reason about reduced order mathematical models based on the Zero Moment Point (ZMP) concept, such as the linear inverted pendulum model \cite{kajita2001}. Nevertheless, these techniques restrict the robot to operate under a small region of its dynamics, where the assumptions of the simplified models are still valid \cite{collins2005,muniz2016}. 

Recent works takes advantages of model-free approaches based on Machine Learning to evade such restrictions and exploit robot dynamics, in order to obtain faster locomotion skills. They are able to achieve running motions \cite{magmafcp2019, fcpfcp2019} and therefore opened a new research direction in the Soccer 3D environment.

In this work, we extend the methodology proposed by \cite{fcprun2019} to learn a new running policy that surpassed the state-of-the-art sprint velocity reported in RoboCup 3D Soccer environment by a significant margin. We obtain this policy through model-free reinforcement learning with no prior knowledge, using a policy gradient algorithm called Proximal Policy Optimization (PPO) \cite{DBLP:journals/corr/SchulmanWDRK17} and features that represents robot's dynamics. Additionally, the results show that the methodology is able to learn policies that surpasses previous state-of-the-art sprint speed in just few hours and much less time than the aforementioned approach.

The remaining of this work is organized as follows. Section \ref{sec:relatedwork} presents related work. Section \ref{sec:background} provides theoretical background. In Section \ref{sec:methodology}, we explain the methodology used in this work. Furthermore, Section \ref{sec:results_and_discussion} presents simulation results to validate our approach. Finally, Section \ref{sec:conclusion} concludes and shares our ideas for future work.

\section{Related Work}\label{sec:relatedwork}

 Many works have experimented on using machine learning and optimization algorithms to develop fast and stable walking motions. \cite{tgmaximo} proposed a walk engine based on periodic functions for the joints trajectories and optimized its parameters via Particle Swarm Optimization (PSO). \cite{AAMAS11-urieli} compared the performance of several algorithms to optimize individual skills in Soccer 3D environment, such as Genetic Algorithms, Hill Climbing, Cross-Entropy Method and Covariance Matrix Adaptation Evolution Strategy (CMA-ES). \cite{AAAI12-MacAlpine} extended the previous work by designing an omnidirectional humanoid walk and by optimizing it using CMA-ES. This work is considered a ``winning approach" in the RoboCup competition that year and it is especially valuable due to the succesfull application of Layered Learning \cite{Stone:1997:LLM:1867406.1867557} to optimize multiples subtasks.
 
 In terms of model-free learning in the context of RoboCup 3D Soccer Simulation League, \cite{Dorer2017LearningTU} use Genetic Algorithms to learn behaviors in joint space.\cite{mcalpine2017} reported a method to optimize keyframe motions using TRPO algorithm. This work has been extended by \cite{tgluck}, which proposes a learning framework that firstly imitates the motion in a neural network \cite{DBLP:journals/corr/abs-1901-00270} and then optimizes it using PPO algorithm.
 
 Finally, in terms of the running motion, \cite{magmafcp2019} applied a modified version of the method presented in \cite{Dorer2017LearningTU} to learn a running behavior from scratch using the toe joints. The state-of-the-art running skill (in terms of forward velocity inside Soccer 3D environment) has been achieved by \cite{fcprun2019}, which also learns a running policy from scratch, but using PPO algorithm and features regarding the robot's dynamics as observed state. 
 
 Comparing our work with the last described, we also use PPO algorithm, the same action space and sprint optimization task. However, in constrast to it, our approach use a more complete state space, and change PPO's hyperparameters and how the policy roll-outs were collected. We also do not perform any modification in the server to perform learning. Our methodology is able to surpass the best velocity reported in \cite{fcprun2019} with significant margin, while reducing the sample complexity to surpass it.

\section{Background}
\label{sec:background}

\subsection{Markov Decision Processes}
We address policy learning in continuous action spaces. We consider the problem of learning a running motion as a Markov Decision Process (MDP), defined by the tuple $M = (\mathcal{S}, \mathcal{A}, \mathcal{P}, r, \rho_{0}, \gamma, T)$, in which $\mathcal{S}$ is a state space, $\mathcal{A}$ is an action space, $\mathcal{P}: \mathcal{S} \times \mathcal{A} \times \mathcal{S} \rightarrow \mathcal{R}_{+}$ a transition probability distribution, $r: \mathcal{S} \times \mathcal{A} \rightarrow [-r_{bound}, +r_{bound}]$ a bounded reward function, $\rho_{0} : \mathcal{S} \rightarrow \mathcal{R}_{+}$ an initial state distribution, $\gamma \in [0, 1]$ a discount factor and $T$ the length of the finite horizon.

During policy optimization, we typically optimize a policy $\pi_{\boldsymbol{\theta}} : \mathcal{S} \times \mathcal{A} \rightarrow \mathcal{R}_{+}$, parameterized by $\boldsymbol{\theta}$, with the objective of maximizing the cumulative reward throughout the episode:

\begin{equation}
\max_{\boldsymbol{\theta}} \mathbb{E}_{\tau}\Big[\sum_{t=0}^{T} \gamma^{t} r(s_{t}, a_{t})\Big],
\end{equation}
where $\tau$ denotes the trajectory, $s_{0} \sim \rho_{0}(s_{0})$, $a_{t} \sim \pi_{\boldsymbol{\theta}}(a_{t} \mid s_{t}$, and $s_{t+1} \sim \mathcal{P}(s_{t+1} \mid s_{t}, a_{t})$.

\subsection{Policy Gradients}
In Policy Gradient (PG) methods, the objective is to learn a parameterized policy that directly select an action from the state space. During training, such methods compute an estimate of the policy gradient and uses into a stochastic gradient ascent algorithm. If we consider cumulative reward as our objective function, we can derive the following Equation for estimating the gradient \cite{DBLP:journals/corr/SchulmanWDRK17}:

\begin{equation}
    \hat{g} = \mathbb{E}_{t} \big[ \nabla_{\boldsymbol{\theta}} \log \pi_{\boldsymbol{\theta}} (a_{t} \mid s_{t})\hat{A}_{t} \big],
\end{equation}
where $\hat{A}_{t}$ corresponds to an estimator of the advantage function. Therefore, in PG algorithms we basically collect some roll-outs from the current policy, estimate an advantage function using the received rewards and then estimate a policy gradient w.r.t the parameters $\boldsymbol{\theta}$, and finally updates such parameters.

The major advantage is that the method is completely model-free, i.e, the gradient itself does not depend on the dynamics. Nevertheless, applying such gradient directly will not result in a good policy, because such estimation is very noisy, resulting in catastrophic updates that slow learning \cite{Sutton1998}.

\subsection{Proximal Policy Optimization}

Proximal Policy Optimization is a family of PG methods for reinforcement learning, which alternate between sampling data through interaction with the environment, and optimizing a “surrogate” objective function using stochastic gradient ascent \cite{DBLP:journals/corr/SchulmanWDRK17}. It takes some benefits of trust region optimization in terms of reliability and stability by defining a ``clipped" surrogate objective:

\begin{equation}
    \mathcal{L}(\boldsymbol{\theta}) = \mathbb{E}_{t}\Big[ \min(r_{t}(\boldsymbol{\theta})\hat{A}_{t}, \text{clip}(r_{t}(\boldsymbol{\theta}), 1 - \epsilon, 1 + \epsilon)\hat{A}_{t}) \Big], 
\end{equation}
where $\epsilon$ is a clip hyperparameter, and $r_{t}(\boldsymbol{\theta})$ is the probability ration defined in Equation \ref{eq:probratio}:

\begin{equation}\label{eq:probratio}
    r_{t}(\boldsymbol{\theta}) = \frac{\pi_{\boldsymbol{\theta} (a_{t} \mid s_{t})}}{\pi_{\boldsymbol{\theta}_{old} (a_{t} \mid s_{t})}}.
\end{equation}

In this way, the clip function avoids excessively large policy updates and reduces the problem of catastrophic steps. 

We use an actor-critic style of PPO, where we also predict the value function and use it to estimate the advantage function through Generalized Advantage Estimation (GAE) algorithm \cite{Schulmanetal_ICLR2016}. Finally, we use a implementation for PPO \cite{baselines} that collects data from multiple parallel actors and synchronize them by applying an average of computed gradients into an unified policy representation (neural network).

\section{Methodology}
\label{sec:methodology}

In this section, we provide details about our methodology: the formulation of running motion as an MDP; the description of the optimization tasks used do obtain our final policy and how we evaluate it; and the configuration regarding the PPO training.

\subsection{Domain Description}

The RoboCup 3D simulation environment is based on SimSpark \cite{10.1007/978-3-662-44468-9_59}, a generic physical multi-agent system simulator. SimSpark uses the Open Dynamics Engine (ODE) library for its realistic simulation of rigid body dynamics with collision detection and friction. The Nao robot has height of approximately 57 cm and 4.5 kilograms. 

The agent sends speed commands to the simulator and receives perceptual data. Each robot has 22 joints with perceptors and effectors, and the monitoring/control of such joints happens at each cycle (20 ms). Visual information is obtained by the agent in periods of 60 ms through noisy measurements of the distance and
angle to objects within a restricted vision cone of 120 degrees. The agent also receives noisy data from sensors: gyroscope, accelerometer and feet pressure. Communication between agent and server happens in the frequency of 50 Hz \cite{LNAI12-MacAlpine2}.

\subsection{MDP Description}

The objective is to obtain a policy that provides actions to a simulated humanoid robot inside RoboCup 3D Soccer environment, in order to run as fast as possible. To achieve this, we modeled the state space with the features reported in Table \ref{tab:statespace}. We aim that, through optimization, the policy learns about the dynamics to perform running with enough stability by using such low-level information.

\begin{table}[htbp]
\caption{State Space}
\begin{center}
\begin{tabular}{|p{2cm}|p{5cm}|p{0.5cm}|}
\hline
\textbf{Feature}&{\textbf{Description}}& \textbf{Size} \\
\hline
Joints' Values & Nao Joints, except from neck yaw and pitch & 20  \\
\hline
General Counter & A counter that increments at each time step & 1 \\
\hline
Left/Right Foot Counter & A counter that restarts when the left/right foot touches the ground and increments at each time step & 2 \\
\hline
Torso's Height and Orientation & The height (relative to ground) and yaw orientation of Nao's torso at the moment & 2 \\
\hline
Center of Mass & The coordinates of each time step & 3 \\
\hline
Torso's Velocity & Torso's coordinates provided by the gyroscope sensor & 3 \\
\hline
Torso's Acceleration & Torso's own angular acceleration coordinates provided by the accelerometer sensor & 3 \\
\hline
Left/Right Foot Pressure Data & The force and origin coordinates computed by left/foot pressure feet sensors & 12 \\
\hline
Rate of Change & The rate of change w.r.t to last time step of each feature previously described, except to the counters & 43 \\
\hline
\end{tabular}
\label{tab:statespace}
\end{center}
\end{table}

As we want to model the running motion in the joint space, give the actual joints' values as input is straightforward. The general counter has the purpose of explicit the sequential nature of decision making to the policy, and helps to improve velocity accordingly to \cite{fcprun2019}. The foot counters, on the other side, explicit the motion period of each leg, which can be useful to maintain symmetry. 

Torso's height, orientation, velocity, acceleration, and center of mass position aim to provide useful information regarding robot's kinematics, as well as the feet force data. Finally, the rate of change is obtained by numeric differentiation and gives important information about the past to complete the observation.

In terms of action space, we use the same indirect approach described by \cite{fcprun2019}. We firstly bound the neural network's output to the interval $[-1, 1]$. We then linearly project  this space onto the joint space, considering the range of possible values to each joint. Finally, we use these target angles to compute the angular velocity of each actuator, using a proportional controller with constant $k = 7$. We saturate the velocity of each joint using the limits provided in Simspark's documentation \cite{simsparkdoc}. 

\begin{figure}[!htbp]
\centering
\includegraphics[width=0.5\textwidth]{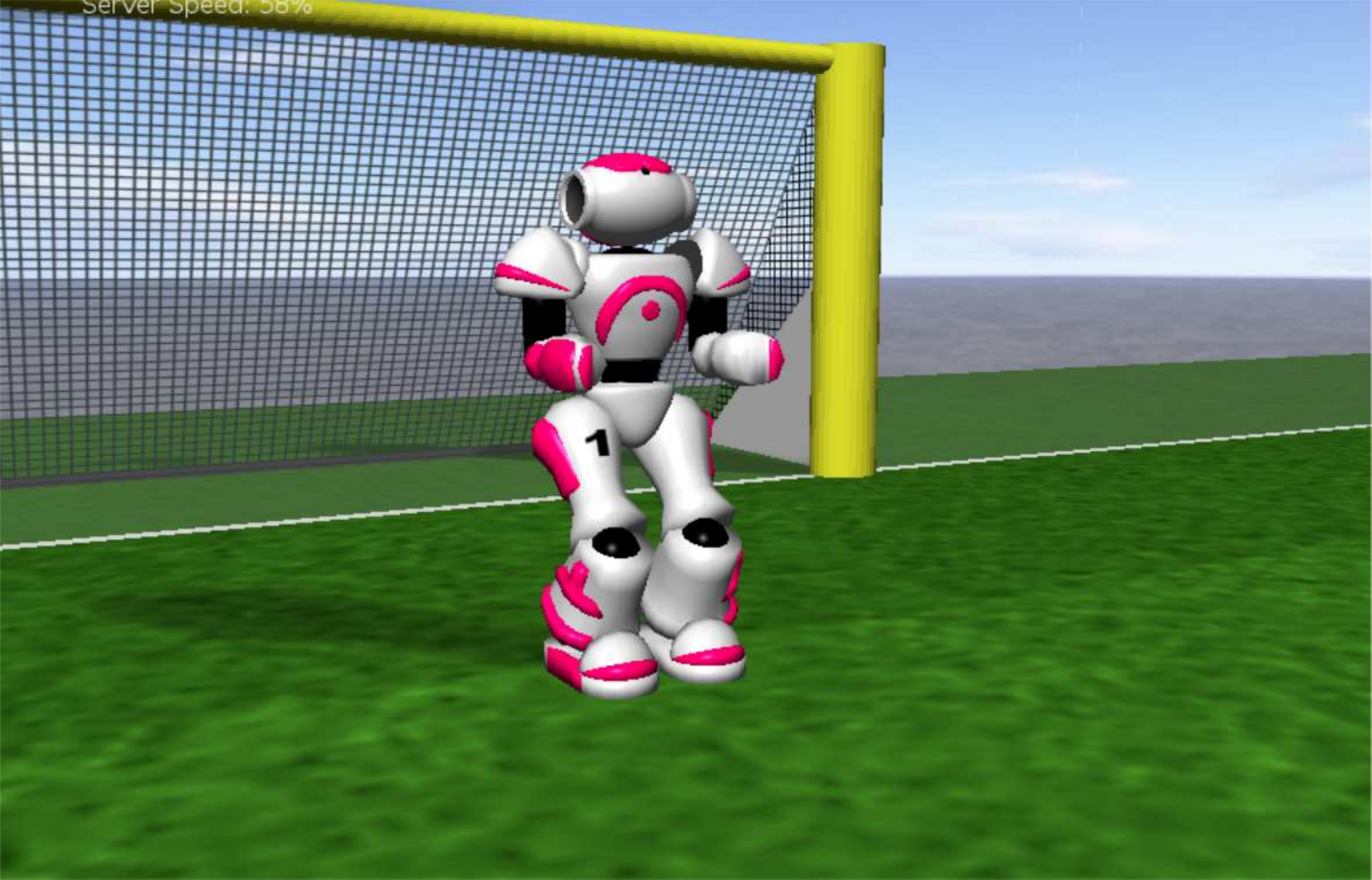}
\caption{Initial robot's joints configuration. }
\label{fig:initialjoints}
\end{figure}

The reward function and episode horizon will depend on the optimization task used and will be described in next section. In terms of initial state distribution, we considered the initial robot's joints configuration that enables the robot to start upright and easily explore bipedal balance and locomotion (Figure \ref{fig:initialjoints}).

\subsection{Optimization Task and Evaluation}

To achieve a sprint motion that run as fast as possible, we create two similar optimization tasks. In both cases, we started the robot at $(-14, 0)$ using the initial joints' configuration previously described. The reward is just the forward distance traveled w.r.t the last time step. 

In the Task I, the policy does not have any prior knowledge, thus we use Early Termination \cite{peng2018} by finishing the episode when the robot falls. This technique helps in two ways: first, it avoids to collect data from a bad terminal state, which the robot is not able to recover itself; second, we explicitly reinforce the agent to keep going forward as long as possible, obtaining more reward. Additionally, we also finish this task when the robot reaches the finish line placed at $x = 14$, which avoids that the agent crashes into the goal post.

The Task II is very similar to the first one, but we consider a fixed episode length of 400 time steps instead of a finish line. In the first task, when the policy is able to achieve the finish line without fall, it starts trying to obtain reward by improving its forward velocity, but it also reduces the episode length and therefore the cumulative reward. A fixed horizon, on the other hand, will avoid this trade-off. We also maintained the Early Termination in case of agent fall.

We evaluate policies by two factors. First, we measure how fast the robot can run by computing its forward velocity. Secondly, how reliable and stable is such locomotion skill, by using the information about angle deviation from a target line. The results reported in next section will use such metrics during and after training.

\subsection{Hyperparameters and Training Procedure}

We used a modified version of PPO's implementation from OpenAI Baselines \cite{baselines}, whose code is available in github\footnote{\url{https://github.com/alexandremuzio/baselines/tree/neural-engine-dynamics}}. Specifically, we used the MPI implementation, which allows parallel agents through MPI processes. We trained using Intel DevCloud \cite{inteldevcloud}, a cluster of Intel Xeon scalable processors. As we have 20 available computation nodes, we used 19 agents collecting data and a master node running the reinforcement learning algorithm.

We ran each optimization task during 200M time steps, which lasts approximately 20 hours in the hardware setup described. We used the hyperparameters from Table \ref{tab:hypers}. Nevertheless, we did not try many sets. As described in \cite{mscluck}, PPO is very sensitive to such parameters, thus we consider their optimization as future work.

Both actor and critic networks use the default architecture implemented in \cite{baselines}: fully-connected networks with two hidden layers of 64 neurons and $tanh$ activation. Weights are initialized as a gaussian distribution with unit variance.

\begin{table}[htbp]
\caption{PPO Hyperparameters}
\begin{center}
\begin{tabular}{|c|c|}
\hline
\textbf{Hyperparameter}&\textbf{Value}\\
\hline
Timesteps per actorbatch & 4096\\
Clip parameter & 0.1 \\
Entropy Coefficient & 0.0  \\
Optimization epochs & 10 \\
Learning rate & 0.0001 \\
Batch size & 64 \\
Discount factor & 0.99 \\
GAE $\lambda$ & 0.95 \\
Learning rate decay & No decay \\
\hline
\end{tabular}
\label{tab:hypers}
\end{center}
\end{table}

\section{Results and Discussion}\label{sec:results_and_discussion}

In this section, we present the results regarding our methodology during training and evaluation, in the light of the metrics previously described. 

In terms of reproducibility, we open source all the training logs (in Tensorboard \cite{tensorflow2015-whitepaper} format), evaluation data and scripts that computed the following results, as well as the trained models\footnote{\url{https://drive.google.com/open?id=1wDEWSQv48qEM8Q17ydPQsrLM7sbtxtwz}}. We also present some videos to illustrate the locomotion skill.

Although we are not able to release the whole agent code (due to competition reasons), we released the portion that corresponds to the training agent, which details the whole MDP implementation \footnote{\url{https://github.com/luckeciano/humanoid-run-ppo}}.

\subsection{Training Procedure}

\begin{figure}[!htbp]
    \centering
    \begin{subfigure}[b]{0.4\textwidth}
        \centering
        \includegraphics[width=1.2\textwidth]{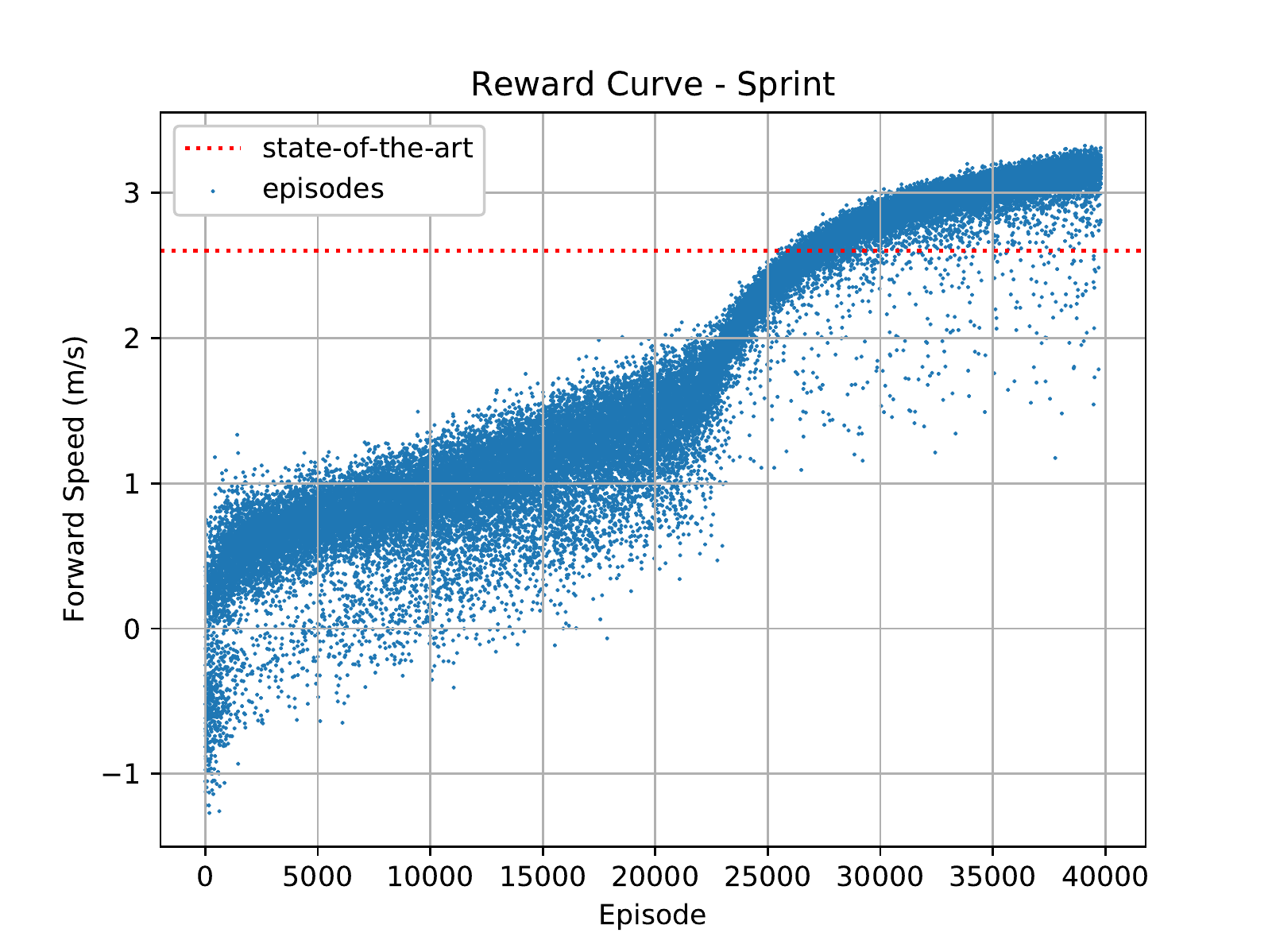}
        \caption{Reward Curve from Sprint Task I}
        \label{fig:rw1}
    \end{subfigure}
    \begin{subfigure}[b]{0.4\textwidth}
        \centering
        \includegraphics[width=1.2\textwidth]{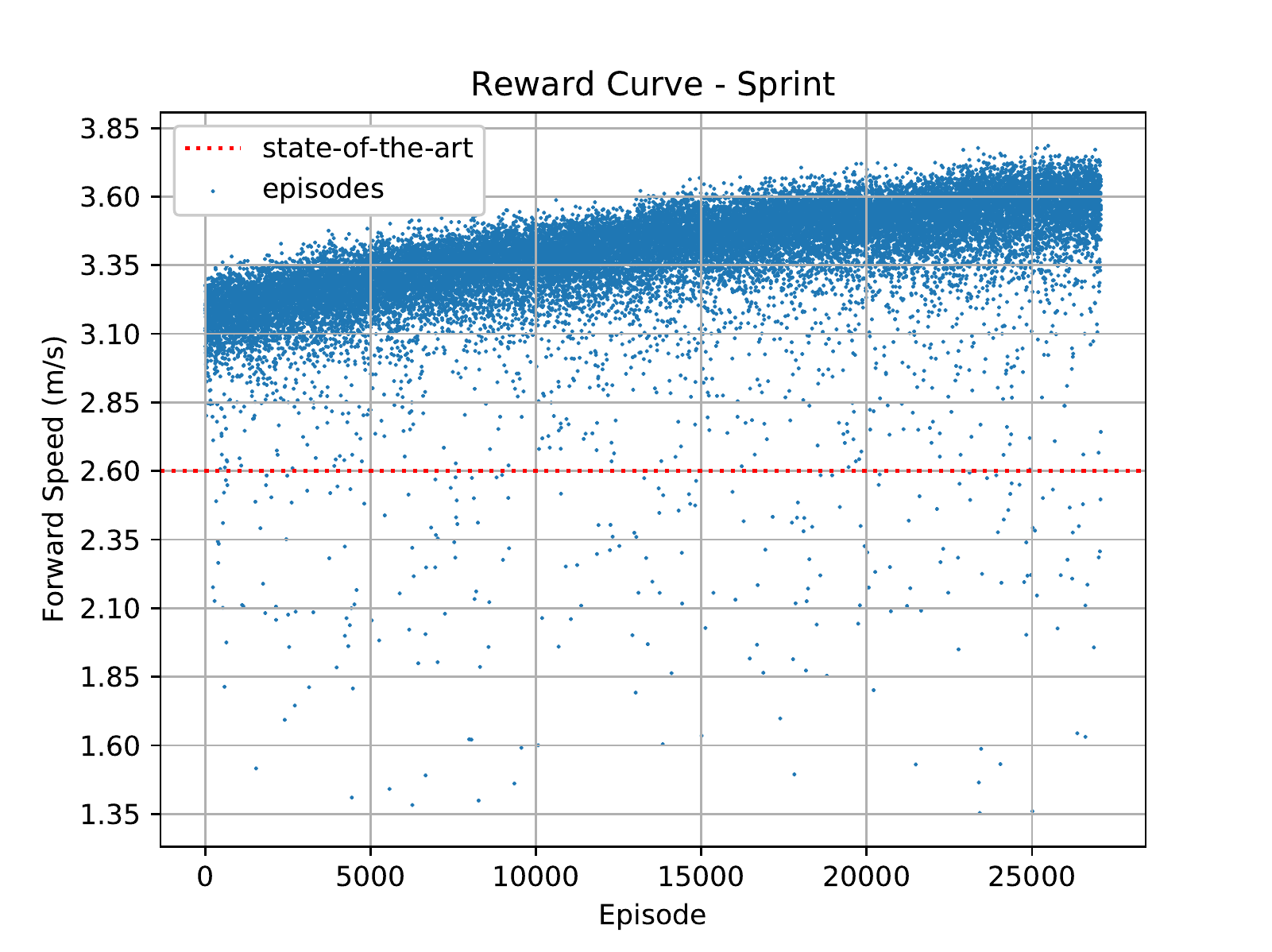}
        \caption{Reward Curve from Sprint Task II}
        \label{fig:rw2}
    \end{subfigure}
\end{figure}

Figures \ref{fig:rw1} and \ref{fig:rw2} present the reward curves from both training procedures, each of them with 200M time steps. These data were collected using one training actor. We also highlight the state-of-the-art forward speed reported inside Soccer 3D environment.

Using the training setup previously described, we use approximately 20.5 and 18.5 hours for training tasks I and II, respectively. The first task achieved the previous best speed between episodes 25000 and 30000, which corresponds to approximately 72M time steps. This shows an improvement, in terms of sample efficiency, considering the results reported in \cite{fcprun2019} (i.e, we reduce the number of samples needed to achieve the same performance). We also observed that approximately 4 hours of training is enough to the agent cross the whole soccer field.

\subsection{Speed Evaluation}

\begin{figure}[!htbp]
\centering
\includegraphics[width=0.5\textwidth]{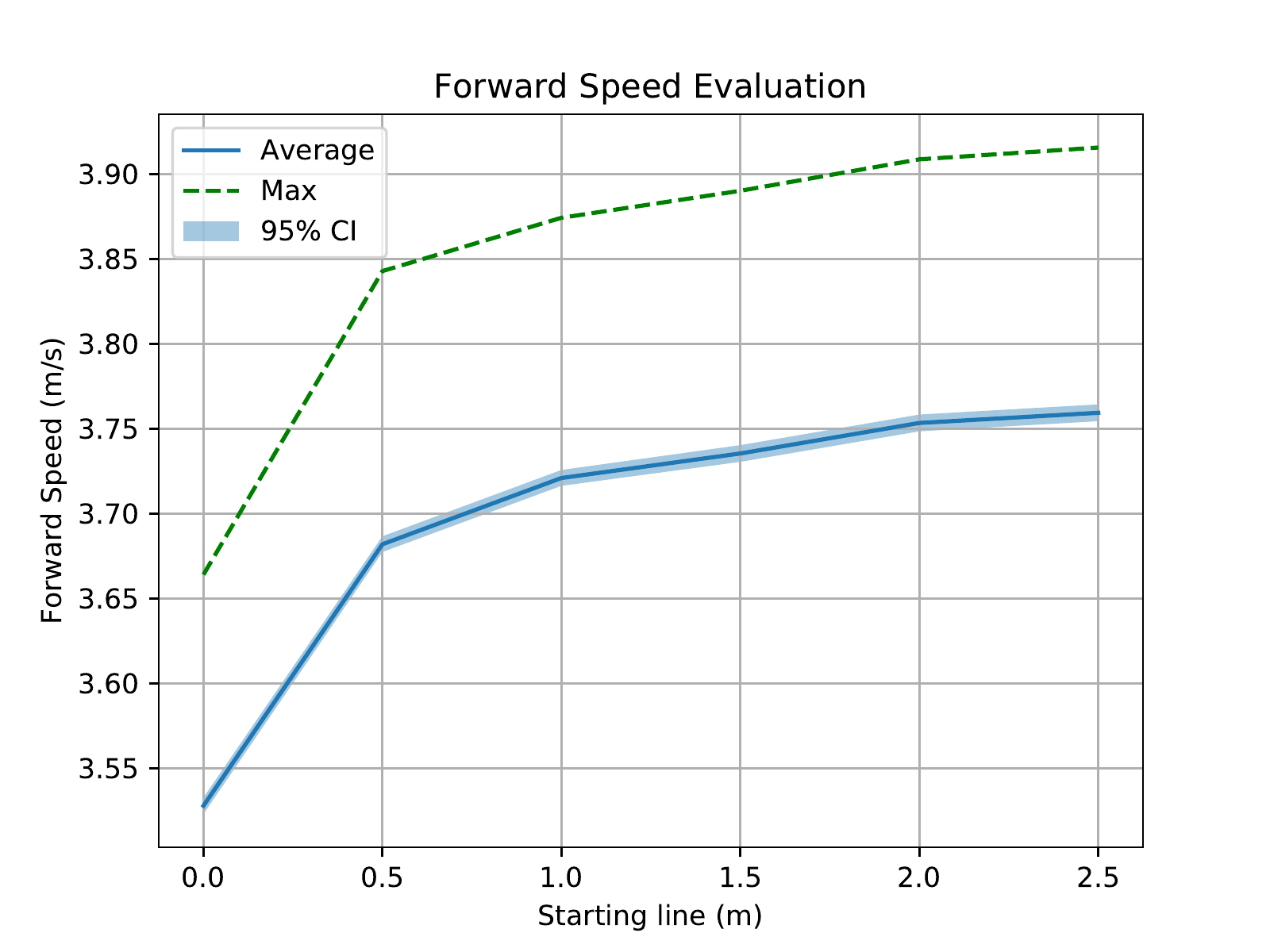}
\caption{Forward speed evaluation from a given starting line. All statistics were collected from 1000 episodes. }
\label{fig:speedeval}
\end{figure}

Figure \ref{fig:speedeval} presents the data regarding speed evaluation. We collected them by reproducing the running motion during 1000 episodes of sprint task I, using the deterministic policy after both training tasks. We present the average and maximum velocities across all episodes. Finally, we also show the 95\% bootstrap confidence interval, symbolized by the blue shaded area.

Accordingly to Figure \ref{fig:speedeval}, we report a top speed of 3.91 m/s, which surpass the best velocity reported in Soccer 3D environment by approximately 50.3\%. The standard deviation for this top speed is 0.07 m/s. Furthermore, we observe a small confidence interval, which reinforces the reliability of the metric presented.

\subsection{Reliability and Robustness}

We also present results about the reliability and robustness of the running motion. We evaluate them by plotting the followed trajectories and evaluating the final deviation.

\begin{figure}[!htbp]
\centering
\includegraphics[width=0.5\textwidth]{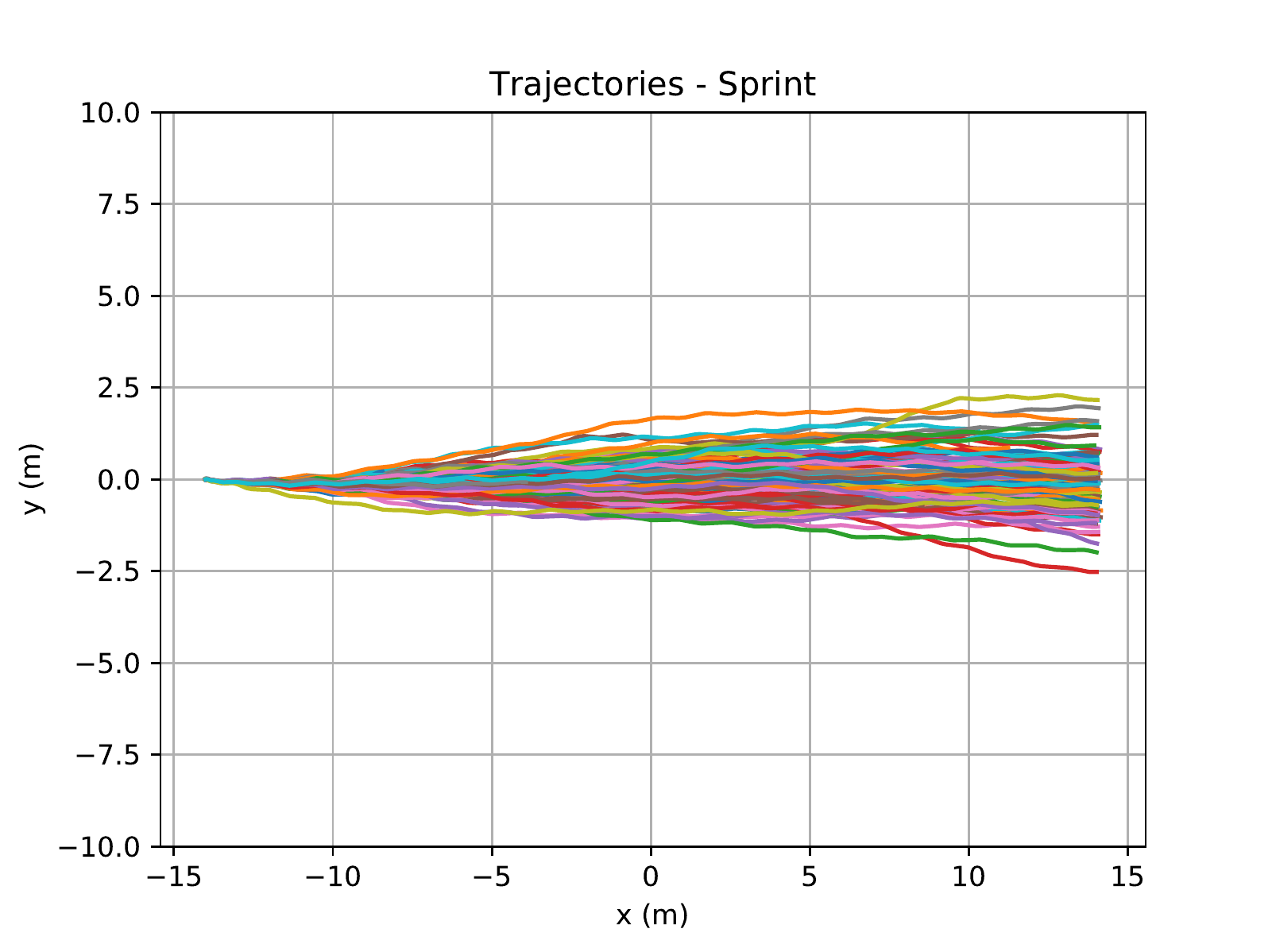}
\caption{Plot from the trajectories followed by the agent in 100 episodes. }
\label{fig:traj}
\end{figure}

Figure \ref{fig:traj} shows the trajectories followed by the agent in 100 episodes. We preferred not to plot all 1000 episodes for the sake of readability. Nevertheless, we report the mean of final deviation across all 1000 collected episodes: 1.52 degrees (from the x-axis), with standard deviation of 1.27 degrees. We did not employ any compensation in agent's pose in order to reduce this deviation. 

In the worst cases presented in the Figure \ref{fig:traj}, there is a deviation of approximately 2.5 meters, that we do not conceive as harmful considering the lenght of the trajectory and the game conditions in RoboCup 3D Soccer Simulation environment. Furthermore, we consider that such deviation can be reduced by applying compensation in the agent's pose input.

\subsection{Human Similarity}

Finally, we need to present qualitatively ideas about how similar the running motion is in comparison to humans. As previously stated, we released videos about the motion\footnote{\url{https://youtu.be/FLkVNh_I3UA}}.

We observed that, although the running motion has fast locomotion skills, the agent's torso is not completely erect, being less human-like. It is intrinsically related to the constrain in place in robot torso's height.

\begin{figure*}[!htbp]
\centering
\includegraphics[width=1.0\textwidth]{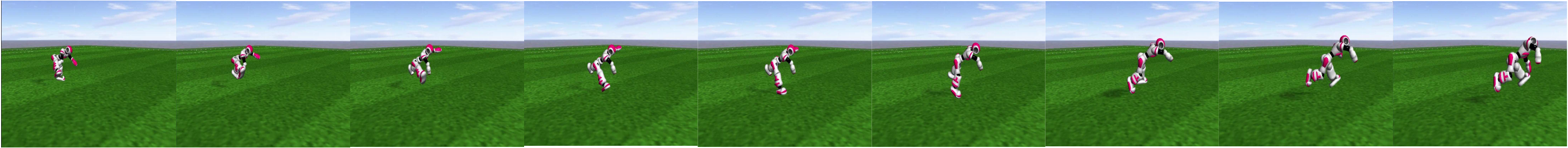}
\caption{Sequential frames illustrating the running motion from the best reported results, in terms of forward speed.}
\label{fig:bestmotion}
\end{figure*}

\begin{figure*}[!htbp]
\centering
\includegraphics[width=1.0\textwidth]{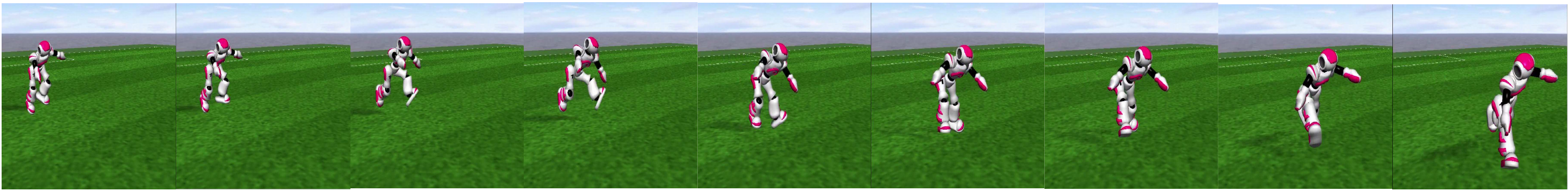}
\caption{Sequential frames illustrating the erect running motion, which is more similar to human locomotion.}
\label{fig:erectmotion}
\end{figure*}

We then reproduced all training procedures previously described, but constraining the minimum robot torso's height to 0.33 m (in contrast to 0.27 m). It resulted in a more human-like motion, at the cost of some stability (the robot falls in more episodes) and forward velocity (top speed of 3.81 m/s) . Figures \ref{fig:bestmotion} and \ref{fig:erectmotion} present both motions as sequences of frames. We also released the all data and plots from both motions, to provide further comparison between them.

\section{Conclusion and Future Work}
\label{sec:conclusion}
In this work, we presented a methodology based on Deep Reinforcement Learning that learns running skills without prior knowledge. We applied the Proximal Policy Optimization algorithm to learn a neural network policy whose inputs are related to the robot's dynamics. The results shows this method is able to surpass the top forward speed by approximately 50.3\%, considering the previous best results reported in \cite{fcprun2019}. Additionally, it is able to learn the motion in few hours, which demonstrates improvements regarding sample efficiency.

During our experiments, we highlight some key factors that are very important to obtain good policies:

\begin{itemize}
    \item We found that torso's height and center of mass are very important to speedup training and obtain faster motions;
    \item Using many parallel actors improved gradient estimation, which avoids bad steps during training;
    \item PPO is very sensitive to its hyperparameters. We tested few sets and each one leads to very distinct policies; and
    \item The training is also very sensitive to some agent's hyperparameters, such as the minimum torso's height and the constant from the proportional controller. 
\end{itemize}{}

Finally, as future work, we plan to apply a curriculum approach to obtain high level behaviors that emerges from this running policy, such as navigation and conduct skills. We also plan to learning from scratch other interdependent skills, like kick and get up motions.

\section{Acknowledgements}
We thank our general sponsors Altium, ITAEx, Mathworks, Metinjo, Micropress, Polimold, Rapid, Solidworks, ST Microelectronics, Wildlife Studios, and Virtual Pyxis.
We specially would like to acknowledge Intel for providing all the computational resources and specialized AI software needed to execute this research.

Finally, we are also grateful to ITA and all the ITAndroids team, especially Soccer 3D simulation team members for the hard work in the development of the base code.

\bibliographystyle{plain}
\bibliography{references}

\end{document}